\pdfoutput=1

\documentclass[11pt]{article}

\usepackage{latex/acl}

\usepackage{times}
\usepackage{latexsym}
\usepackage{graphicx}
\usepackage{amsmath}
\usepackage{comment}
\usepackage{caption}
\usepackage{subcaption}
\usepackage{makecell}
\usepackage{booktabs}
\usepackage{multirow}
\usepackage{multicol}
\usepackage[T1]{fontenc}

\usepackage[utf8]{inputenc}

\usepackage{microtype}

\usepackage{inconsolata}

\title{Identifying Multiple Personalities in Large Language Models \\with External Evaluation}

\author{\textbf{Xiaoyang Song$^1$, Yuta Adachi$^{2*}$, Jessie Feng$^{2*}$, Mouwei Lin$^{2*}$, Linhao Yu$^{2*}$, Frank Li$^{2*}$}, \\ \textbf{Akshat Gupta$^3$, Gopala Anumanchipalli$^3$, Simerjot Kaur$^4$}\\
$^1$University of Michigan, $^2$Columbia University, $^3$UC Berkeley, $^4$ AI Research, JPMorgan \\
 \texttt{xysong@umich.edu, akshat.gupta@berkeley.edu, simerjot.kaur@jpmchase.com}}

\begin{document}
\maketitle
\begin{abstract}
As Large Language Models (LLMs) are integrated with human daily applications rapidly, many societal and ethical concerns are raised regarding the behavior of LLMs. One of the ways to comprehend LLMs' behavior is to analyze their personalities. Many recent studies quantify LLMs' personalities using self-assessment tests that are created for humans. Yet many critiques question the applicability and reliability of these self-assessment tests when applied to LLMs. In this paper, we investigate LLM personalities using an alternate personality measurement method, which we refer to as the \textit{external evaluation} method, where instead of prompting LLMs with multiple-choice questions in the Likert scale, we evaluate LLMs' personalities by analyzing their responses toward open-ended situational questions using an external machine learning model. We first fine-tuned a Llama2-7B model as the MBTI personality predictor that outperforms the state-of-the-art models as the tool to analyze LLMs' responses. Then, we prompt the LLMs with situational questions and ask them to generate \textit{Twitter posts} and \textit{comments}, respectively, in order to assess their personalities when playing two different roles. Using the external personality evaluation method, we identify that the obtained personality types for LLMs are significantly different when generating posts versus comments, whereas humans show a consistent personality profile in these two different situations. This shows that LLMs can exhibit different personalities based on different scenarios, thus highlighting a fundamental difference between personality in LLMs and humans. With our work, we call for a re-evaluation of personality definition and measurement in LLMs.
\end{abstract}

\section{Introduction}\label{sec:Introduction}
\footnote{$^*$ equal contribution}
The evolution of Large Language Models (LLM) has benefited humans in the past few years through their unprecedented capacities to understand and generate human-like languages \citep{gpt1, gpt2, gpt3, gpt3.5, gpt4, chatgpt, opt, touvron2023llama}. For instance, LLMs are now being deployed as virtual assistants to provide mental health support \citep{lai2023psy}, as online educators for common knowledge retrieval \citep{chatgpt, llmeducation}, and even as helpers in symbolic music compositions \citep{agostinelli2023musiclm, imasato2023using}. However, this growing integration of LLMs across different social sectors of human life raises important concerns about reliability, safety, and ethics. This dual nature of LLMs opens the need to study their behaviors, especially their behaviors when interacting with humans. Although most chat-based models including ChatGPT \citep{chatgpt} and Llama \citep{llama2} have undergone safety training to prevent delivering poisonous and biased information, there is still an urgent need to find a proper venue and metrics to understand their societal behaviors.

One common way to understand the behavior of LLMs is to measure their personalities through rigorous psychometric studies \citep{personality2-mpi, personality3-whoisgpt, personality5-enfj, personality4-identifying, personality1-karra}. According to the American Psychological Association (APA), personality for humans is defined as \textit{``the enduring characteristic and behavior that comprise a person's unique adjustment to life"} \citep{personality_definition-apa}. On the other hand, the exact definition of LLM personalities remains an open yet mysterious question in the field. Nevertheless, many researchers made analogies to human personality and attempted to study LLM personalities by leveraging the psychometric tests used for humans. For instance, most recent literature prompted LLMs with standardized self-assessment personality test questions and then recorded and analyzed the results \citep{jiang2023evaluating, personality1-karra, personality3-whoisgpt, personality6-temporal, personality7-berkeley}. However, while these tests are shown to be effective for human personality measurement \citep{psych_personality-1}, there is evidence that they can not be directly applied to reliably measure personality of both base LLMs including GPTs \citep{gpt1, gpt2,gpt3} and chat-based LLMs like ChatGPT \citep{chatgpt} and Llama \citep{llama2}. For instance, \citet{personality0-ours} and \citet{gupta2023investigating} managed to show that the self-assessment personality test results of the same LLM differ significantly as the prompting templates are changed, which is not surprising as LLMs are known to be prone and sensitive to different prompts \citep{sclar2023quantifying, chen2023unleashing}. However, it has also been shown that even under the same prompt template, the psychometric test results are statistically different when the options of those multiple-choice questions are presented in different orders \citep{personality0-ours, gupta2023investigating}. These observations indicate that prompting LLMs with standardized self-assessment is not a reliable method to quantify LLMs' personalities. Therefore, an alternative to self-assessment psychometric tests is desired to better understand and analyze personalities in LLMs.

In this paper, we investigate an alternate method to measure LLM personalities. To do so, we first develop a state-of-the-art personality prediction model. Specifically, we utilized the famous Myers-Briggs Type Indicator (MBTI) personality framework and fine-tuned a Llama2-7B \citep{llama2} model on a human personality dataset. The dataset contains multiple posts written by a human subject and their MBTI personality type. This model is then used to analyze the personality of LLMs under the MBTI framework and compare the results with human counterparts. (Sec. \ref{sec: stage-i})



To measure LLM personality, we have different LLMs write tweets which are used as input for our personality prediction model. While doing this, we have LLMs take two different roles. In the first role, the LLM is asked to write \textit{tweets} about real-world events based on the event topics which were obtained by analyzing news articles. This is done to prevent data leakage and stop LLMs from repeating tweets seen previously in pre-training data. In the second role, we ask the LLM to write \textit{replies} to existing tweets. The tweets are again collected in real time and are not part of the model pre-training corpus. We then evaluate the personality of different LLMs based on the tweets generated by them using our personality prediction model. We perform this analysis for ChatGPT \cite{chatgpt}, Llama2-7B-chat, Llama2-13B-chat and Llama2-70B-chat models \cite{llama2}. At the same time, we also repeat the same procedure for human-written posts and comments to validate the proposed personality detection model. To our surprise, we find that the personality distribution of LLMs when writing tweets is completely different from the distribution when responding to tweets. As defined by APA, personality is supposed to be an \textit{enduring characteristic} for humans. While this consistency is shown to be true for humans, we show that LLMs exhibit different personalities while playing different roles. (Sec. \ref{sec: stage-ii} \& Sec. \ref{sec: stage-iii}). 

To summarize, our paper makes the following contributions:
\begin{itemize}
    \item We fine-tune a Llama2-7B for MBTI personality detection that significantly outperforms the state-of-the-art models.
    \item We use the fine-tuned personality detection model on tweets made by humans and show that human personality remains consistent across different roles.
    \item We show that LLMs exhibit different personalities across different roles when using the external evaluation method.
    \item With our work, we call for a re-evaluation of personality definition and measurement in LLMs.
\end{itemize}

\section{Related Work}
\subsection{Personality Theory}
Human personality is defined as \textit{``the enduring characteristic and behavior that comprise a person's unique adjustment to life"} by APA \citep{personality_definition-apa}, which is typically measured across different effective trait dimensions \citep{personality-cattel1, personality-cattel2}. Central to human psychological profiling are the \textit{Big Five} personality traits, also referred to as the OCEAN traits, which stands for \textbf{O}penness, \textbf{C}onscientiousness, \textbf{E}xtraversion, \textbf{A}greeableness, and \textbf{N}euroticism, respectively \citep{psych_personality-1, psych_personality-2, psych_personality-3}. The \textit{Big Five} personality traits are typically measured through multiple choice self-assessment questions where a situational statement is presented and the test-takers are requested to choose from an option in Likert scales, typically 1 to 5, to reflect the degree of fitness of the statement to themselves \citep{ipip-neo-120}. Many studies on LLM personality directly base their works on this straightforward test \citep{jiang2023evaluating, personality0-ours, personality4-identifying}.

\begin{figure}[t!]
    \centering
    \includegraphics[width=0.45\textwidth]{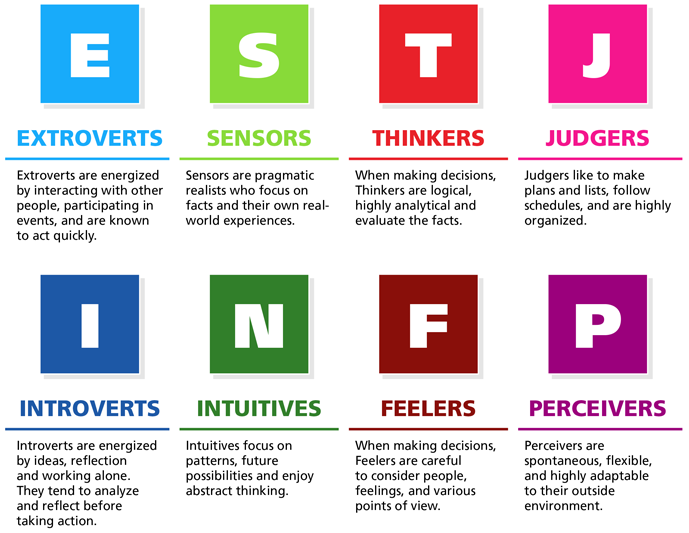}
    \caption{MBTI trait description \citep{mbti-fig}.}
    \label{fig:MBTI}
\end{figure}

However, the result of this test is a distribution of scores among all five traits, making it difficult to analyze and compare. Instead, we build our paper on another famous categorical personality framework called the Myers-Briggs Type Indicator (MBTI), which details 16 distinct types based on trait combinations across four different dimensions: \textbf{E}xtraverts vs. \textbf{I}ntroverts (E/I), \textbf{S}ensors vs. \textbf{I}ntuitives (S/I), \textbf{T}hinkers vs. \textbf{F}eelers (T/F), and \textbf{J}udegers vs. \textbf{P}erceivers (J/P). Figure ~\ref{fig:MBTI} above provides detailed information about the trait each dimension measures.

\subsection{LLM Personality Measurement}
Many recent works regarding LLM personality asked the LLM to perform Multiple-Choice Question-Answering (MCQA) on those well-known standardized self-assessment personality tests \citep{jiang2023evaluating, personality3-whoisgpt, personality4-identifying, personality5-enfj, personality6-temporal, personality7-berkeley, personality8-mbti, personality9-text2behavior}. For instance, as one of the foundation works, \citet{jiang2023evaluating} prompted the LLMs with the widely-used IPIP-120 dataset in psychology \citep{ipip-neo-120} and evaluated their \textit{Big Five} scores (i.e. OCEAN scores). However, although these works managed to elaborate on how personality can play a role in LLMs, the validity of these self-assessment tests remains unchecked, making their conclusions unreliable. As shown by \citet{personality0-ours} and \citet{gupta2023investigating}, the same LLM tends to exhibit significantly different personalities on different self-assessment tests, which fails to achieve the crucial \textit{consistency} criterion in personality definition and highlights the invalidity of self-assessment tests. The origins of these concerns are related to both the difficulty of MCQA tasks and the fact that LLMs are sensitive to the change in prompts as well as the structure of the prompts. A well-known example of this is that the LLMs can be manipulated logically via Chain-of-Thoughts (CoT) \citep{wei2022chain}.

To the best of our knowledge, there is not much literature that attempts to evaluate LLM personality without leveraging standardized tests. Driven by this gap, in this work, we attempt to completely discard self-assessment tests and use an external personality assessment method, where a personality detection model is used to predict LLM personality. In addition, unlike previous works that conduct tests without validation, we also justify the personality prediction model used for evaluation by conducting a validation experiment on humans.

\subsection{MBTI Personality Detection Models}
Most MBTI personality detection models are built based on the public Kaggle dataset\footnote{https://www.kaggle.com/datasets/datasnaek/mbti-type} \citep{tang2023attention, yang2023orders, mehta2023personality}. The dataset consists of over 8600 entries, where each entry corresponds to an individual's MBTI type and includes excerpts from the last 50 posts and comments made by the individual on the PersonalityCafe forum\footnote{https://www.personalitycafe.com/forums/myers-briggs-forum.49/}. Detailed introduction to example entry, label distribution, and the tweet topic distributions can be found in Appendix \ref{apd: kaggle}. 

To predict MBTI personality types from texts, \citet{mehta2023personality} processed the texts using a BERT model \citep{devlin2018bert} and then trained an MLP to predict the MBTI personality type, while \citet{yang2023orders} utilized a graph convolutional neural network to learn the connections between different posts made by an individual in order to make decent predictions. Furthermore, \citet{tang2023attention} proposed an attention-based denoising framework (ADF) for MBTI personality detection, where they trained the model to effectively extract personality signals from noisy and verbose text data. However, a common weakness in these works is that their methods predict the personality type for each MBTI dimension separately. Therefore, although their methods achieve an average of around $70\%$ prediction accuracy for each trait dimension, the overall performance remains mysterious when the predictions for each dimension are aggregated together because the errors and uncertainties accumulate multiplicatively. In our method, we fine-tuned the popular Llama2-7B \citep{llama2} model as the predictor, which outperformed their models significantly.

\label{sec: experiments}
\begin{figure*}[t]
     \centering
     \includegraphics[scale=0.67]{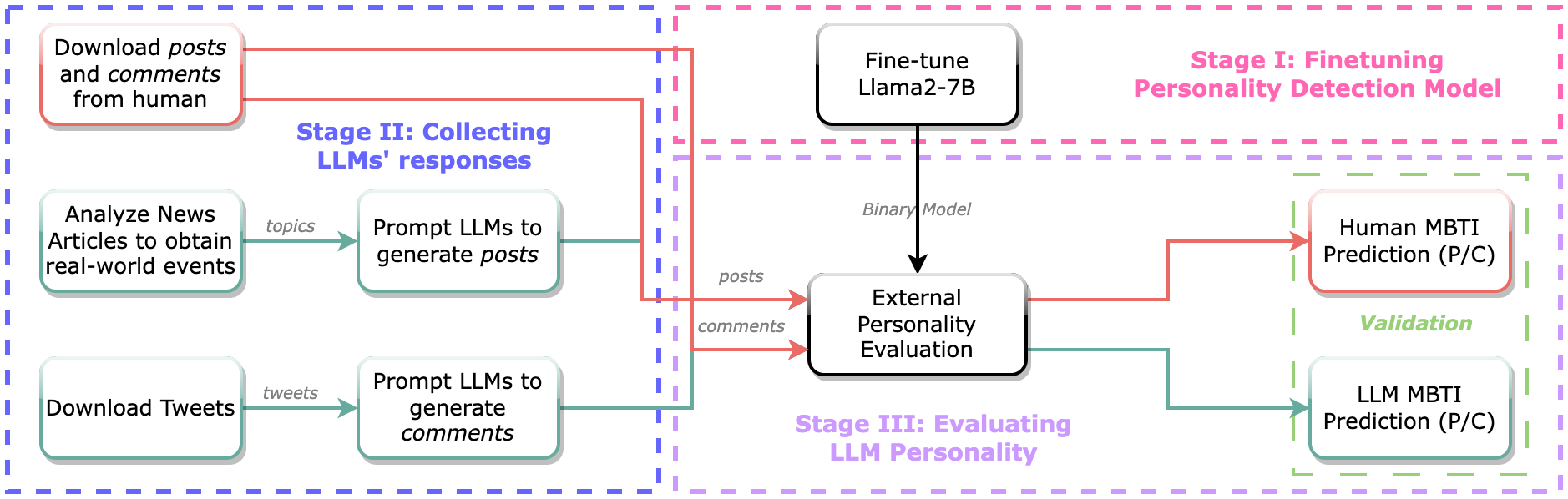}
     \hfill
     \caption{Methodology flowchart.}
     \label{fig: flowchart}
\end{figure*}

\section{Experiments}
In this paper, we refer to using personality prediction model to measure personality as an \textit{external personality evaluation method}, juxtaposing it with the personality self-assessment methods. The experiments on the proposed external evaluation method are divided into three stages. Firstly, a Llama2-7B \citep{llama2} model is fine-tuned on the public MBTI datasets as the personality detection model. Then, to perform external evaluations, we prompted the LLMs with pre-processed situational questions/scenarios and prompted them for open-ended generations based on the input. Finally, the collected responses from the LLMs were fed into the personality detection model to obtain the evaluation results. In this paper, we base our experiments on four popular chat-based LLMs: ChatGPT \citep{chatgpt} and three versions of Llama2 (7B, 13B \& 70B) \citep{llama2}. The overall experiment pipeline can be found in Figure \ref{fig: flowchart}. In addition, the computational resources used are introduced in Appendix \ref{apd: comp-resources}.

\subsection{Stage I: Fine-tuning Llama2-7B-based Personality Detection Model}
\label{sec: stage-i}

In this work, we utilized the widely used aforementioned Kaggle MBTI dataset to fine-tune the personality detection model. Although many previous studies build their models on the same dataset \citep{yang2023orders, tang2023attention, mehta2023personality}, there is room for improvement in the overall performance of their models, which may affect the reliability of personality prediction.

In particular, we proposed two models: (1) a binary model and (2) a 16-class model. In terms of the binary model, we fine-tuned four pre-trained Llama2-7B models, and each of them is designed to be a binary classifier for each of the four MBTI dimensions. The predictions are then aggregated to form the final prediction, which is similar to most of the previous works \citep{tang2023attention, yang2023orders, mehta2023personality}. On the other hand, in the 16-classes model, a pre-trained Llama2-7B is directly fine-tuned to predict one of the sixteen classes. Aligning with the Kaggle dataset format, both models take 50 posts from an individual as inputs at a time to predict personality. The fine-tuning process was conducted under the LoRA \citep{hu2021lora} framework, where the target modules to fine-tune are the query (q) and the value (v) layers. We chose the rank to be $r=16$ and fine-tuned it for only 5 epochs with a learning rate of $10^{-4}$ and batch size of 8. The training started with a warm-up phase for the first 100 iterations, followed by a linear learning rate decay.

\begin{table}[h!]
    \centering
    \scalebox{0.71}{
    \begin{tabular}{|c|c|c|c|c|} \hline 
         Models&  Accuracy &  F1 score &  Precision & Recall \\ \hline 
         \makecell[c]{BERT-Base + MLP\\ \citep{mehta2023personality}} &  73.1&  63.2 &  78.0 & 65.5 \\ \hline 
         \makecell[c]{D-DGCN\\ \citep{yang2023orders}} &  78.2&  53.5 &  60.5 & 49.1\\ \hline 
         \makecell[c]{ADF\\\citep{tang2023attention}}&  61.0&  38.8&  47.5& 56.8\\ \hline 
         \makecell[c]{\textbf{Llama2-7B + FT (B)}\\{\textbf{(ours)}}}&  93.3 &  91.1 &  \textbf{92.3} & 90.1\\ \hline
         \makecell[c]{\textbf{Llama2-7B + FT (16)}\\{\textbf{(ours)}}}&  \textbf{93.5} &  \textbf{91.4} &  92.0 & \textbf{90.8}\\ \hline
    \end{tabular}
        }
    \caption{Average performance of personality detection models on each of the four MBTI dimensions. Each cell is a numerical average of the performance of each binary model in percentage.}
    \label{tab:ft}
\end{table}

\begin{table}[h!]
    \centering
    \scalebox{0.72}{
    \begin{tabular}{|c|c|c|c|c|} \hline 
         Models&  Accuracy&  F1 score &  Precision & Recall \\ \hline 
         \makecell[c]{BERT-Base + MLP\\ \citep{mehta2023personality}} &  29.4 & 8.70 & 14.1 & 12.0 \\ \hline 
         \makecell[c]{D-DGCN\\ \citep{yang2023orders}} & 37.6 &  35.9 & 35.9 & 37.6\\ \hline 
         \makecell[c]{ADF\\\citep{tang2023attention}}&  14.0 &  4.10 &  3.28 & 5.74\\ \hline 
         \makecell[c]{\textbf{Llama2-7B + FT (B)}\\{\textbf{(ours)}}}&  81.0 & 74.3 & 76.9 & 73.1\\ \hline
         \makecell[c]{\textbf{Llama2-7B + FT (16)}\\{\textbf{(ours)}}}&  \textbf{81.7} &  \textbf{76.9} &  \textbf{79.8} & \textbf{75.2}\\ \hline
    \end{tabular}
        }
    \caption{Performance of personality detection models on 16-class MBTI prediction tasks. The numbers in the cell are all percentages.}
    \label{tab:ft-overall}
\end{table}

The testing performance of the proposed personality detection model and the baselines are reported. With the capacities of LLMs to understand texts and extract key knowledge, both the fine-tuned binary model and 16-classes model outperform the state-of-the-art models significantly in all metrics. In Table \ref{tab:ft}, we report the average performance of the model for each trait dimension. To do this, the accuracy of the model in predicting each of the four dimensions of MBTI personality is calculated separately. Then the accuracy for these four dimensions is averaged to present the numbers in Table \ref{tab:ft}, which is the most common form of evaluation used in prior work \citep{tang2023attention, yang2023orders}.

\begin{table*}[t!]
\centering
\scalebox{0.85}{
\begin{tabular}{p{2cm}| p{5cm} | p{10cm}}
\hline
\textbf{Type}  & \multicolumn{1}{c|}{\textbf{System Prompt Used}} & \multicolumn{1}{c}{\textbf{User Prompt Used}}\\
\hline
Posts & \texttt{Generate a Twitter post} & \texttt{As a user on Twitter, write a tweet on the following contents: [summarized contents]}\\\hline
Comments & \texttt{Generate a Twitter comment} & 
 \texttt{As a user on Twitter, write a tweet to comment on this Tweet: [tweet contents]}\\\hline
\end{tabular}}
\caption{Prompting templates.}
\label{tab: template}
\end{table*}

One drawback of such an evaluation is that it doesn't take into account the overall accuracy of the model in predicting one of the 16 classes of personality. In Table \ref{tab:ft-overall}, we take into account the overall prediction errors and present the aggregated 16-class classification performance. The difference in performance between Table \ref{tab:ft} and Table \ref{tab:ft-overall} for the same models highlights the effectiveness of our model. For instance, while the baseline methods achieve around $70\%$ accuracy when averaged over each trait, when those predictions are aggregated together to predict the actual 16-classes MBTI type, the accuracy drops significantly to around $30\%$. However, for our fine-tuned binary model, the prediction accuracy for each trait is $93.3\%$ while the overall accuracy remains as high as $81.0\%$. This evaluation method helps highlight error accumulation during the personality evaluation procedure, rendering the experimental results more accurate. We observed that the binary model and 16-classes model yield comparable performance for nearly all the metrics. In later experiments, we chose to stick to the binary model as it is often used in previous works \citep{tang2023attention, yang2023orders, mehta2023personality}, making the results comparable. Complete results for fine-tuned models can be found in Appendix \ref{apd:ft}.

\begin{figure*}[t]
     \centering
     \includegraphics[scale=0.45]{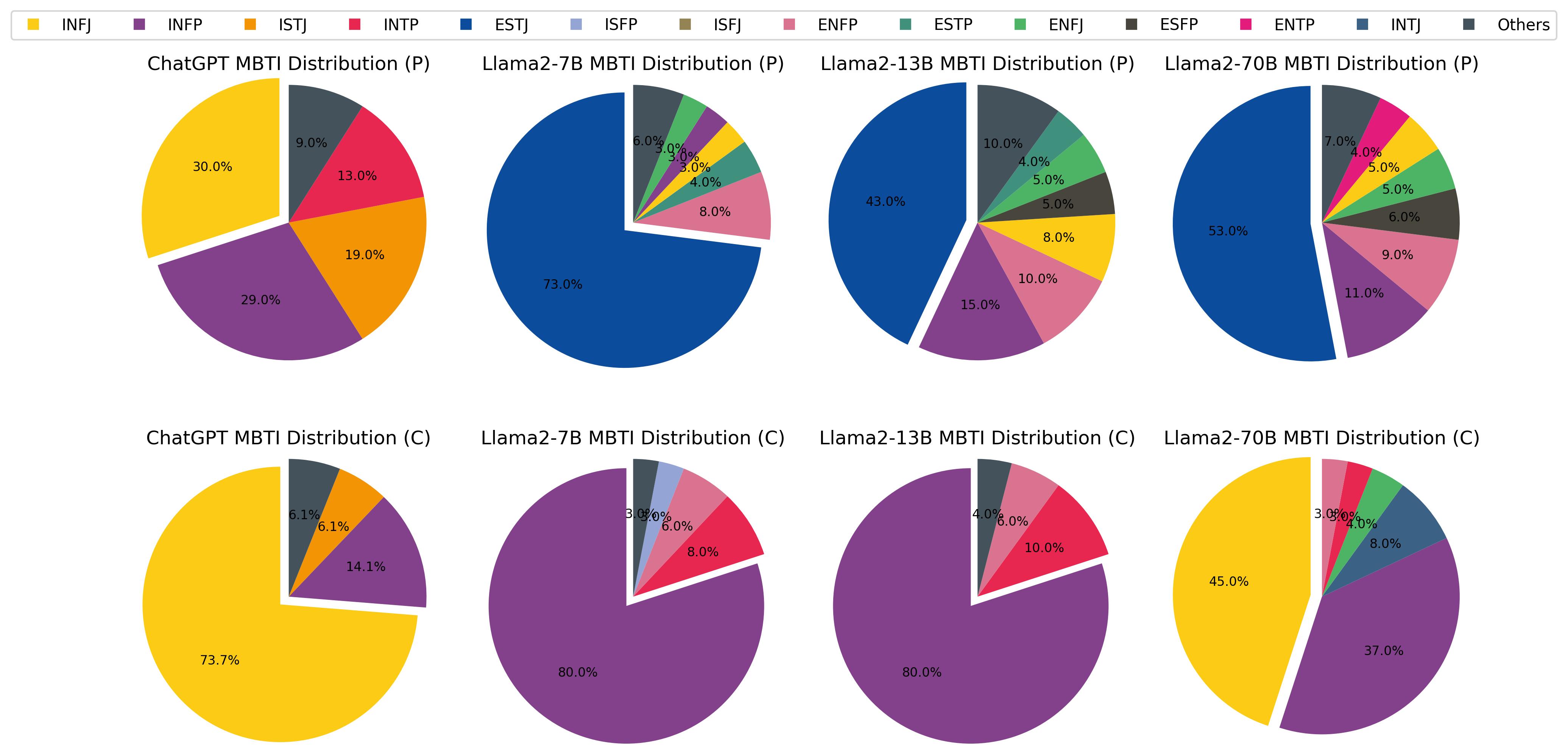}
     \hfill
     \caption{MBTI distribution of ChatGPT and Llama2 models using external evaluation method for 100 times. In the figure, the first row is the assessment results on the generated posts dataset (P), whereas the second row provides the results on the comments dataset (C). In addition, personality types that appear less than 3 times are merged together to form the class ``Others''.}
     \label{fig: pie}
\end{figure*}

\subsection{Stage II: Collecting LLMs Responses Toward Open-ended Situational Questions}
\label{sec: stage-ii}
In the second stage, we prompt LLMs with open-ended situational questions and collect their responses as inputs for external evaluation. This is done to collect social media post data generated by an LLM which can be used as input for our personality prediction model. 


\noindent\textbf{Prompting Pipeline. } To effectively analyze daily news, prompt LLMs, and process their responses, we designed a prompting pipeline. The pipeline starts by analyzing news from different topics and summarizing them into the latest news events before prompting the LLMs. We analyzed the latest news articles from November 2023 to make sure that the events were not present in the training data of the personality prediction models. These latest news events serve as the basis for prompting the LLMs to generate tweets, with the LLMs taking on the role of individuals posting tweets about these news events. Additionally, we also gather online tweets written by other individuals and prompt the LLMs to generate comments in response to these tweets. This allows the LLMs to assume different roles, specifically that of replying to tweets made by others. To prevent any potential data leakage, these tweets are collected exclusively from the month of November 2023. 

Depending on whether we want the LLMs to write \textit{posts} (i.e. \textit{tweets}) about the summarized events or \textit{comments} (i.e. \textit{replies}) to the existing tweets, the prompting templates are chosen to be different accordingly. For instance, the template for prompting LLMs to write a post is the following: ``\textit{As a user on Twitter. Write a tweet on the following contents:} \texttt{[summarized contents]}''. It is noteworthy that we chose not to engineer the prompting templates in order to keep them as simple as possible, so as to not induce behavior of LLMs. The exact prompts used in this work are shown in Table \ref{tab: template}. We use the simplest prompts to elicit generation in order to not bias the model through prompts for generations.

In total, we analyzed several thousand news articles to extract relevant news events and collected 5000 tweets from 10 popular topics. Subsequently, we tasked the LLMs with generating 4500 posts based on the news events and 5000 comments in response to the collected tweets. These generated texts were then utilized to study the personalities of the LLMs using our personality detection model. It is worth noting that the roles assigned to the LLMs varied depending on whether they were asked to generate posts or comments, allowing for a diverse range of responses. A detailed description of this pipeline and downloaded contents can be found in Appendix \ref{apd: etl}.

\begin{figure*}[t]
     \centering
     \includegraphics[scale=0.45]{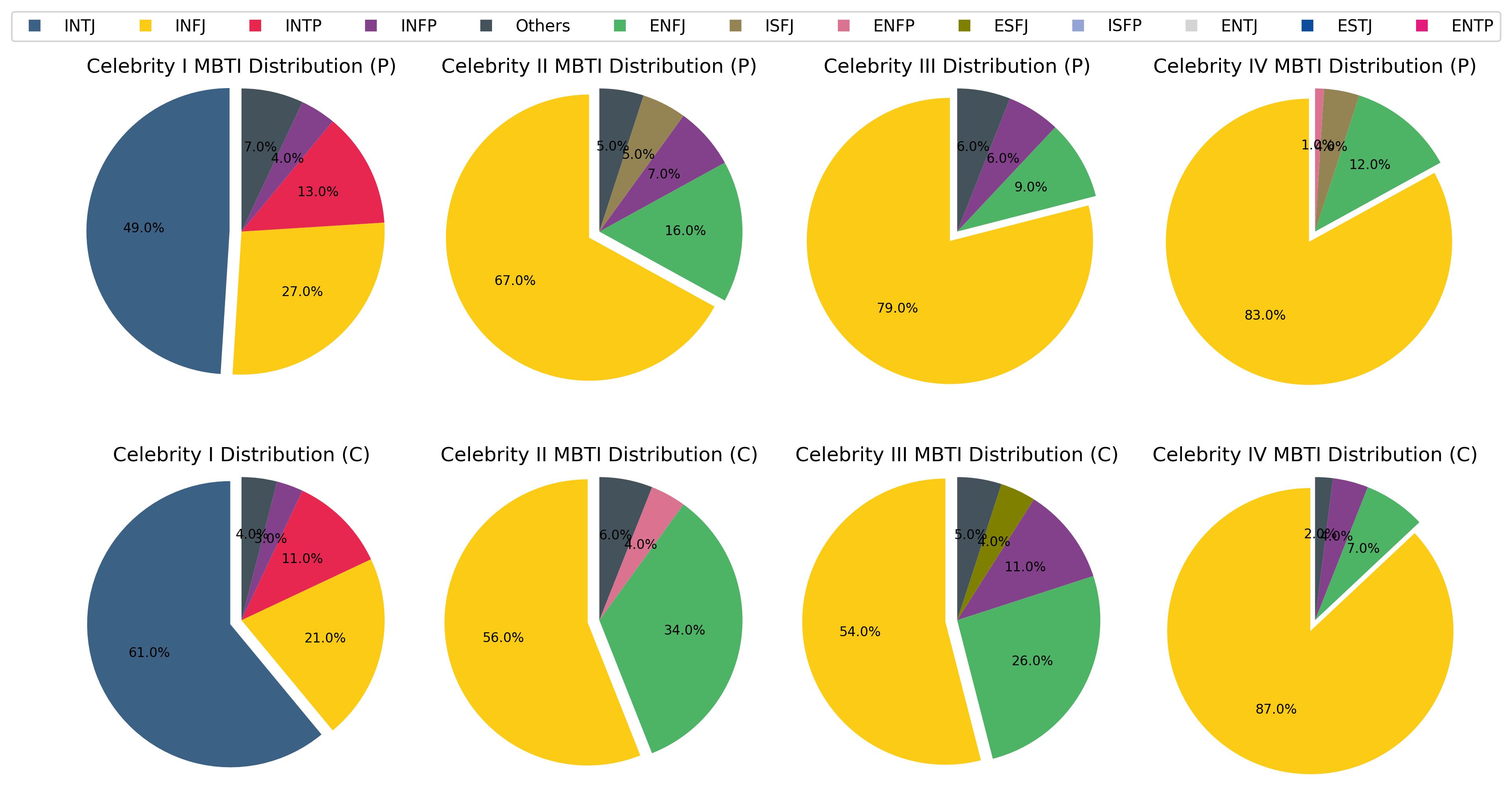}
     \hfill
     \caption{MBTI distribution of 4 celebrities using the external evaluation method for 100 times. In the figure, the first row is the assessment results on the generated posts dataset (P), whereas the second row provides the results on the comments dataset (C). In addition, personality types that appear less than 3 times are merged together to form the class ``Others''. Complete results for all 8 selected celebrities can be found in Appendix \ref{apd: validation}.}
     \label{fig: pie_human_main}
\end{figure*}

\subsection{Stage III: External LLMs Personality Detection \& Validation}
\label{sec: stage-iii}
After obtaining the posts and comments datasets for each LLM of interest, we evaluate their personalities using our fine-tuned personality detection model. The personality detection model is trained to take 50 social media posts of a user as input and output their personality based on the 16-class MBTI personality framework. To minimize errors due to sampling and model inaccuracy, we created 100 sets of 50 generations for each type of tweet (\textit{posts vs. comments}) generated by the LLM. To do so, we sampled 50 responses with replacements from our generation dataset 100 times. We then report the predicted MBTI personality distribution for the 100 samples for each role in Figure \ref{fig: pie} and the most frequent personality type for each LLM is reported as the predicted personality in Table \ref{tab: result-self}.


\begin{table}[h!]
    \centering
    \scalebox{0.80}{
    \begin{tabular}{|c|c|c|} \hline  
         \multirow{2}{*}{Model} & \multicolumn{2}{c|}{External Evaluation}\\\cline{2-3}
         & Posts & Comments\\\hline
         ChatGPT & INFJ & INFJ \\ \hline  
         Llama2-7B & ESTJ & INFP\\ \hline  
         Llama2-13B & ESTJ & INFP\\ \hline  
         Llama2-70B & ESTJ & INFJ\\ \hline 
    \end{tabular}}
    \caption{MBTI personality type obtained by external evaluation method. The most frequent MBTI personality type is reported here. }
    \label{tab: result-self}
\end{table} 

We can observe from Table \ref{tab: result-self} that for the same LLM, the external evaluation results on the generated posts and comments datasets are very different. For instance, for Llama2-7B, the most frequent MBTI type is ESTJ when it posts tweets, whereas it is INFP on the generated comments. In addition, the predicted personality distributions for the 100 trials in Figure \ref{fig: pie} are also significantly different between posts and comments. For example, in ChatGPT personality evaluation, although the most frequent MBTI type is the same in Table \ref{tab: result-self} between posts and comments, the predicted personality distributions are extremely different, as can be seen in the first column of pie charts in Figure \ref{fig: pie}. We can see that INFJ and INFP are nearly equally distributed when writing posts, whereas INFJ dominates over all other MBTI types when writing comments. In plain words, this indicates ChatGPT \textit{behaves} very differently when generating posts and comments. We can see this very clearly for the other Llama2 models, where the personality distribution while writing posts (first row of pie charts in Figure \ref{fig: pie}) is very different from the personality distribution we get when writing comments (second row in Figure \ref{fig: pie}). \textbf{These experiments clearly show that LLMs exhibit different personalities when playing different roles}.


\noindent\textbf{Validation with Human Counterpart. } Personality is defined as an \textit{enduring characteristic} in humans by the American Psychological Association. This means that for humans, the personality distribution obtained for humans when writing posts versus comments is expected to be consistent. As we employed a personality detection model to evaluate LLM personality, it becomes important to check if the detection model is able to produce similar personality distributions when humans take on the two roles of writing posts versus comments. If this is not the case, then the inconsistencies in LLM personalities in the two roles can be attributed to the personality detection model. Furthermore, this is crucial to justify the correctness of our later analysis, which is completely based on this personality detection model.

To check this, we conduct the same study with human counterparts. We randomly downloaded the \textit{Twitter posts} and \textit{comments} from 8 celebrities\footnote{Due to proprietary reasons, we have masked the identities of the selected celebrities.} from different domains and applied the personality detection model to these two different datasets. We again created 100 samples of 50 tweets each and found the personality distribution for these celebrities in the two different roles - while writing posts versus comments. The personality distribution results for four celebrities can be seen in Figure \ref{fig: pie_human_main}, while the remaining four can be seen in Figure \ref{fig: human_pie}. 

\begin{table}[h!]
    \centering
    \scalebox{0.80}{
    \begin{tabular}{|c|c|c|} \hline  
         \multirow{2}{*}{Celebrity} & \multicolumn{2}{|c|}{External Evaluation} \\ \cline{2-3}
         & P & C \\ \hline  
         Celebrity I & INTJ (INFJ)  & INTJ (INFJ) \\ \hline  
         Celebrity II &  INFJ (ENFJ) & INFJ (ENFJ)\\ \hline  
         Celebrity III &  INFJ (ENFJ) & INFJ (ENFJ)\\ \hline  
         Celebrity IV &  INFJ (ENFJ) & INFJ (ENFJ) \\ \hline
         Celebrity V & INFJ (INFP) & INFJ (INFP) \\ \hline  
         Celebrity VI &  INFJ (INTJ) & ENTJ (ENFJ)\\ \hline  
         Celebrity VII &  INFJ (INFP) & INFJ (INFP)\\ \hline  
         Celebrity VIII &  INFJ (ISFJ) & INFJ (ENFJ)\\ \hline 
    \end{tabular}}
    \caption{The most frequent MBTI personality type for 8 celebrities under external evaluation method using \textit{posts} and \textit{comments}. The second most frequent MBTI personality type is reported in the parenthesis.}
    \label{tab: result-human}
\end{table} 

Table \ref{tab: result-human} presents the results for the eight celebrities. We can see that the most frequent MBTI type, the second most frequent MBTI type, and even the predicted distributions (Figure \ref{fig: pie_human_main}) are very akin to each other for nearly all the celebrities that we investigated. This empirically confirms that humans tend to exhibit \textit{enduring} MBTI personality types even when playing different roles, which echoes the observations in most of the psychology literature \citep{green2019personality, personality_definition-apa} as well as the personality definition \citep{APA:83}. A detailed introduction to this validation experiment can be found in Appendix \ref{apd: validation}. Simultaneously, this confirms that our fine-tuned model is able to attain similar personality distributions for humans and rules out its effects in the observed \textit{inconsistency} in LLMs.

The above experiments show that LLMs clearly exhibit different personalities in different roles, which is not true for humans. Our experiments also show that the definition of personality as defined for humans may not be applicable to LLMs, as personality in LLMs no longer seems to be an enduring characteristic.

\section{Conclusions}\label{sec:Conclusions}
In this paper, we investigate the personality exhibited by LLMs using an \textit{external personality evaluation} method, which is an alternative to self-assessment tests for personality measurement. To perform external personality evaluation, we fine-tuned a Llama2-7B model as the third-party agent for personality detection and collected LLMs' responses and attitudes toward open-ended situational questions. We also developed a prompting pipeline to automate the prompts for LLMs with proper content. Our experiments show that LLMs exhibit different personalities in different roles, which is not true for humans. 

These observations also shed light on the validity of quantifying and defining LLM personality using the same standard for humans. In our work, we see that LLM personality is not an enduring characteristic as it is seen in humans. Additionally, previous works have shown that self-assessment tests are not the appropriate tool to measure personality in LLMs. When we combine our observations with these results, we can conclude that not only do we not have accurate tools to measure LLM personality, but we also lack the appropriate definition of personality for LLMs. With our work, we caution against a naive transfer of definition and methods used to analyze human personality on LLMs and call for more fundamental work on evaluating LLM personality and behavior, taking into account the specific characteristics of LLMs. 



\section{Limitation}
This paper investigates the validity of applying external evaluation to measuring LLM personalities and calls for a re-evaluation of the definition and measurement of LLM personalities. As we employed a fine-tuned model as an external agent to evaluate personalities and based our analysis on it, one of the limitations of our work is the accuracy of the personality detection model. Although our model attains state-of-the-art performance and we reduce the chances of errors by using 100 samples of generations instead of 1, the uncertainties introduced due to error rates are still inevitable. Furthermore, although our work identifies that the current definition of LLM personality should be reevaluated, this paper does not provide an alternative for defining LLM personality and the measurement method, which are left for future work.

\section{Acknowledgement}
This paper was prepared for information purposes by the Artificial Intelligence Research group of JPMorgan Chase \& Co and its affiliates (“JP Morgan”), and is not a product of the Research Department of JP Morgan. J.P. Morgan makes no representation and warranty whatsoever and disclaims all liability for the completeness, accuracy or reliability of the information contained herein. This document is not intended as investment research or investment advice, or a recommendation, offer or solicitation for the purchase or sale of any security, financial instrument, financial product or service, or to be used in any way for evaluating the merits of participating in any transaction, and shall not constitute a solicitation under any jurisdiction or to any person, if such solicitation under such jurisdiction or to such person would be unlawful. © 2023 JP Morgan Chase \& Co. All rights reserved.

\bibliography{anthology,custom}

\newpage
\newpage
\appendix

\section{Appendix}

\subsection{Kaggle MBTI Dataset}
\label{apd: kaggle}
The public Kaggle MBTI dataset is comprised of 8675 entries, each corresponding to an individual's MBTI type. In particular, every data entry includes excerpts from the last 50 posts or comments made by the individual on the PersonalityCafe forum, separated by the special character ``\texttt{$\mid\mid\mid$}''. Two example data entries are provided in Table \ref{tab: example-mbti}. Note that this dataset is anonymous upon creation.

\subsection{Computational Resources}
\label{apd: comp-resources}
In this work, all the experiments are done with Nvidia RTX A6000 GPU. In particular, we hosted Llama2-7B, Llama2-13B, and Llama2-70B on 1, 2, and 4 GPU cards, respectively, where Llama2-70B was hosted in mixed precisions. For the ChatGPT experiment, we directly use OpenAI APIs\footnote{https://platform.openai.com/}.

\subsection{Finetuning Experiments Details}
\label{apd:ft}
In this section, we report the experimental details and complete results for both binary and 16-class models to reproduce Table \ref{tab:ft}. For baseline models, as the models are trained on the same dataset, we kindly refer the audience to the original papers for the performance of their models \citep{yang2023orders, mehta2020bottom, tang2023attention}. In particular, in the fine-tuning experiments, the original MBTI dataset is divided into training, evaluation, and heldout testing subsets in an 81:9:10 ratio. Furthermore, for the binary model, the model for each trait shared the same set of hyperparameters.

\begin{table}[h!]
    \centering
    \scalebox{0.80}{
    \begin{tabular}{|c|c|c|c|c|c|} \hline 
         Metrics &  \textbf{E/I} &  \textbf{N/S} &  \textbf{T/F} & \textbf{P/J} & Average\\ \hline 
          Accuracy ($\%$) &  94.01 &  96.08 &  92.97 & 89.98 & 93.26\\ \hline 
          F1 ($\%$) &  90.57 &  91.48 & 92.91 & 89.38 & 91.09\\ \hline 
          Precision ($\%$) & 93.16 &  93.44 &  93.13 & 89.41 & 92.29\\ \hline 
          Recall ($\%$) & 88.57 &  89.76 &  92.78 & 89.36 & 90.10 \\ \hline 
    \end{tabular}
        }
    \caption{Performance of personality detection models on each of the four MBTI dimensions for the binary model.}
    \label{tab:ft-detail}
\end{table}

\begin{table}[h!]
    \centering
    \scalebox{0.80}{
    \begin{tabular}{|c|c|c|c|c|c|} \hline 
         Metrics &  \textbf{E/I} &  \textbf{N/S} &  \textbf{T/F} & \textbf{P/J} & Average\\ \hline 
          Accuracy ($\%$) &  93.09&96.31&	93.55	&90.90	&93.46\\ \hline 
          F1 ($\%$) &  89.49	&92.10 & 	93.50 & 	90.29	&91.35\\ \hline 
          Precision ($\%$) & 90.33	&93.37	&93.65	&90.64 & 	92.00\\ \hline 
          Recall ($\%$) & 88.71	&90.93&	93.40 &89.99&	90.76 \\ \hline 
    \end{tabular}
        }
    \caption{Performance of personality detection models on each of the four MBTI dimensions for the 16-class model. Note that the numbers are calculated by extracting each dimension from the predictions.}
    \label{tab:ft-detail-16}
\end{table}

\begin{table*}[t!]
\centering
\scalebox{0.85}{
\begin{tabular}{p{3cm} | p{14cm}}
\hline
\textbf{Type}  &\textbf{Contents} (50 posts or comments on PersonalityCafe)\\
\hline
ENTP & I am finding the lack of me in these posts very alarming $\mid\mid\mid$ This + Lack of Balance and Hand-Eye Coordination. $\mid\mid\mid$ ... \\\hline
INTJ & 2\% still means about 1/50 people. I've probably seen 1-2 others today. I never understood fascination by virtue of rarity.$\mid\mid\mid$ I collect shoes. I do so because I like status and nothing communicates such a thing as much as a pair of Jordans... \\\hline
... &...\\\hline
\end{tabular}}
\caption{Two example data entries of the public Kaggle MBTI dataset.}
\label{tab: example-mbti}
\end{table*}

\subsection{Prompting Pipeline \& Downloaded Topics}
\label{apd: etl}

In principle, we want to feed LLMs with real-world topics, and tweets and collect their responses and attitudes toward them. We analyzed daily news to obtain a list of real-world topics.  In terms of the topics for tweets, we downloaded 5000 tweets containing the following 10 different topics: Bitcoin, NFL, Music, Oscars, Travel, Fashion, Food, Fitness, Gaming, and Technology, where they are nearly evenly distributed.

The templates that we used to generate Twitter posts and comments are provided in Table \ref{tab: template}.  As for the parameters for generation, we utilized a nucleus sampling \citep{holtzman2019curious} with the temperature and \texttt{top\_p} set to 0.2 and 0.95, respectively. In addition, the maximum length of generated contents is set to 200. The same configuration is used for all LLMs that we investigated.

\subsection{Validation Experiment on Human Counterpart}
\label{apd: validation}
\begin{figure*}[t]
     \centering
     \includegraphics[scale=0.50]{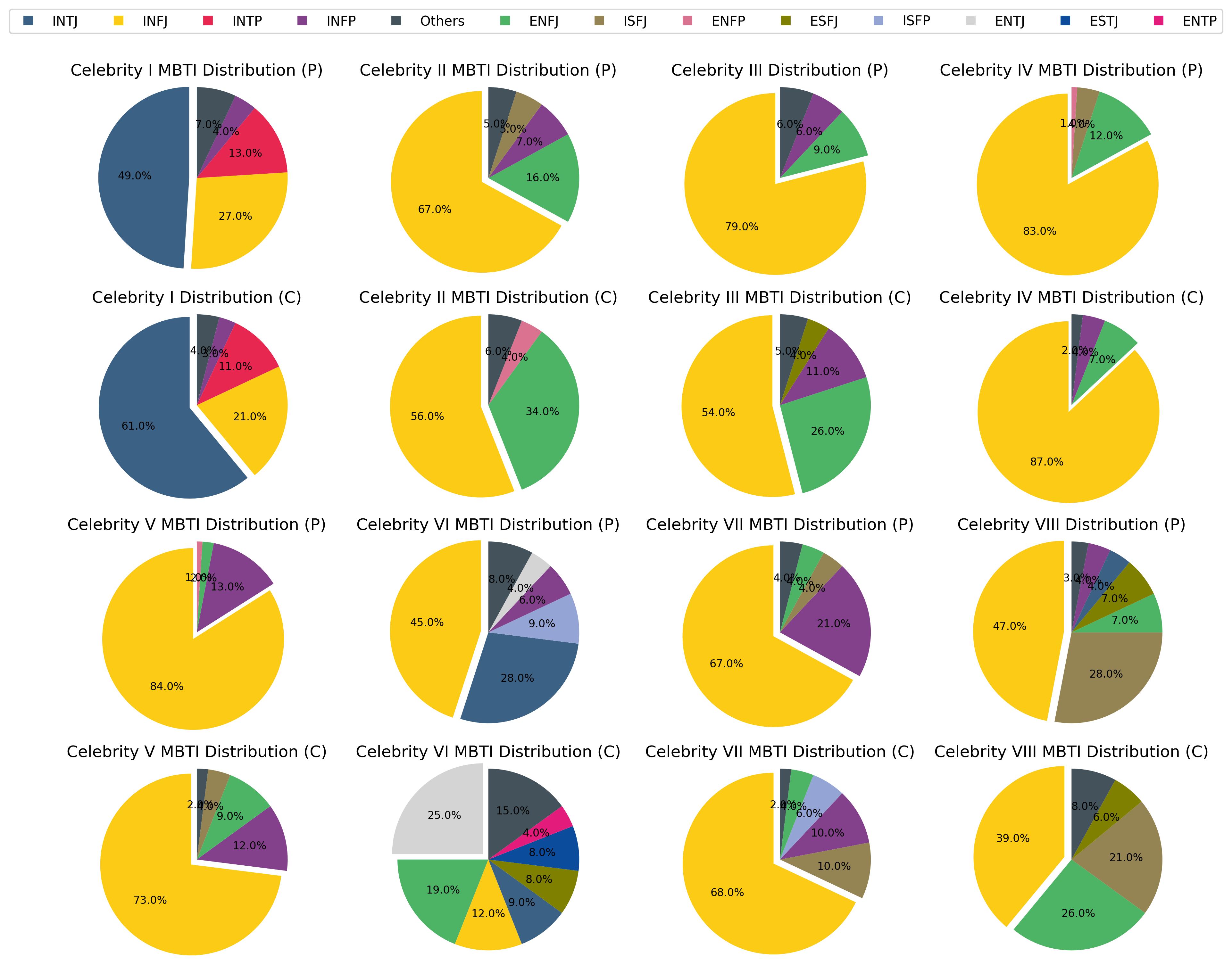}
     \hfill
     \caption{MBTI distribution of 8 selected celebrities using external evaluation method for 100 times. In the figure, the first row is the assessment results on the generated posts dataset (P), whereas the second row provides the results on the comments dataset (C). In addition, personality types that appear less than 3 times are merged together to form the class ``Others''.}
     \label{fig: human_pie}
\end{figure*}

In the validation experiment, we downloaded the public Twitter posts and comments made by 8 celebrities from different domains and use the same personality evaluation procedure as what we have done for LLMs. Specifically, we created 100 samples with 50 tweets in each for every celebrity and repeated the external evaluation method for 100 times. The obtained MBTI distributions and the most frequent MBTI types are reported in Figure \ref{fig: human_pie} and Table \ref{tab: result-human}, respectively, where the results echo the fact in psychology literature where personality is a consistent characteristic for humans \citep{personality_definition-apa, green2019personality}.

\end{document}